\documentclass{article}
 \usepackage{graphicx}
\usepackage{amsthm}
\usepackage{amsmath}
\usepackage{amssymb}
\usepackage{hyperref}

\newcommand{\la} {\langle}
\newcommand{\ra} {\rangle}

\title{Limits of an AI program for solving college math problems} 

\author{
Ernest Davis \\
Dept. of Computer Science \\ New York University \\ New York, NY 10012 \\
{\small davise@cs.nyu.edu}}

\begin{document}
\maketitle

\begin{abstract}
Drori et al. (2022) report that ``A neural network solves, explains, and 
generates university math problems by program synthesis and few-shot learning at
human level ... [It] automatically answers 81\% of university-level
mathematics problems.'' The system they describe is indeed impressive; however,
the above description is very much overstated. The work of solving the problems
is done, not by a neural network, but by the symbolic algebra package Sympy.
Problems of various formats are excluded from consideration. The so-called 
``explanations'' are just rewordings of lines of code. Answers are
marked as correct that are not in the form specified in the problem. Most
seriously, it seems that in many cases the system uses the correct answer given
in the test corpus to guide its path to solving the problem.
\end{abstract}

Drori et al. (2022) report that ``A neural network solves, explains, and 
generates university math problems by program synthesis and few-shot learning at
human level.''  

Specifically, the neural network takes as input word problems from
undergraduate math courses. The wording of the problem is first modified by
a hand-crafted automated front end. In most cases this involves only adding a 
few stock phrases, such as ``Use sympy''; this is called ``zero-shot learning''.
In some cases, the system automatically finds similar examples in the corpus and
adds them to the prompt; this is called ``few-shot learning''.
The
modified problem is given as input to the Codex system developed by OpenAI
(Chen et al. 2021), and Codex outputs Python code intended to solve the problem. 

The system was tested on a data set of
word problems taken from  six MIT undergraduate math courses (Single Variable
Calculus, Multivariable Calculus, Differential Equations, Introduction
to Probability and Statistics, Linear Algebra, and Mathematics for Computer
Science), one Columbia University course (Computational Linear Algebra),
and six topics from the MATH dataset (Hendrycks et al. 2021) (Prealgebra,
Algebra, Number Theory, Counting and Probability, Intermediate Algebra,
and Precalculus), assembled from high school math competitions. Over this
data set, it is claimed the system achieves 71\% success rate using 
zero-shot learning and an additional 10\% success rate using few-shot
learning.

By contrast, the large language model GPT-3 (Brown et al. 2020)
achieves only 18\%
with zero-shot learning and 30.8\% using few-shot learning and
chain-of-thought prompting. Thus the new system is a major advance in the
state of the art.

Unquestionably this is a very impressive accomplishment. However, there are
a number of points that should be kept in mind here in evaluating its 
significance.

First: It is misleading to attribute this success in solving the problems to 
``a neural network'', as in the title. The actual work of finding the 
mathematical solutions is, in all cases, being done by Sympy, a large and 
sophisticated Python package for symbolic mathematics that was constructed by 
hand, building on sixty years of expert development of computational 
symbolic math systems.
What the neural network does is to convert the problems from an English language
statement into the proper call to Sympy functions (the additional 
surrounding Python code is generally trivial.) This itself is of course a 
very difficult problem, and the success of Codex on these problems is remarkable,
but it is important to be clear about which parts of the overall process 
are being done by what kinds of technology.

Second: As mentioned in the paper (p. 2 and p. 9) the program cannot solve 
problems where the question involves an image or requires a proof. The claim
(p. 1) that the system solves ``81\% of university-level mathematics problems''
therefore has to be qualified. (Moreover, of course, there 
are many university-level mathematics problems that go beyond the first six 
introductory courses.)

Third: In a number of the examples given in the paper, the answer provided by
the system is not actually in the form requested in the original problem. For 
instance, Table 1 problem 2 reads ``Describe the graph of the function
$f(x,y) = 10 - \sqrt{x^{2}+y^{2}}$.'' The answer given by the system is this
three-dimensional graph:

\begin{center}
\includegraphics[width=3in]{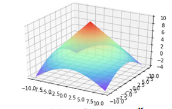}
\end{center}

Presumably, however, the author of the question did not have in mind producing
an image, which many students would find challenging to do by hand, and trivial
to do using plotting software. Rather, the question asks for a verbal
description: 
``The graph is a right circular cone, with apex at $\la 0,0,10 \ra$ and axis
downward on the $z$ axis'' or something of the kind. 

Another example: In figure 4, the first question (from course 18.01) reads
``Find the derivative of the function using the definition of a derivative:
$f(x) = (x**2-1)/(2*x-3)$.'' Presumably the intent of ''using the definition
of a derivative'' is that the student should actually show how the answer
can be derived using the limit definition of a derivative (admittedly a somewhat
laborious problem, requiring a fair amount of rather fiddly algebra). The system,
however, simply ignores the words ``using the definition of a derivative''
and instructs Sympy to use the built-in symbolic differentiation module.

Fourth: The title states that the system ``explains'' the problem. In fact, what 
the system produces is a line-by-line explanation of the Python code that 
computed the answer, generally rather dreary and unenlightening. For instance
in the same example from 18.01 in figure 4 discussed above, the explanation 
produced is as follows
\begin{verbatim}
        1. We import sympy as sp.

        2. We create a symbol x.

        3. We create a function f.

        4. We print the derivative of f with respect to x.
\end{verbatim} 
That really doesn't go far toward explaining the problem. The other examples in
figure 4 are similar.

Fifth, and perhaps the most serious: The way in which problems are 
``automatically'' chosen for few-shot learning is altogether unclear and may
be illegitimate. The paper says (figure 2) ``If zero-shot learning does not
work, perform few-shot learning,'' and (p. 4) ``If the question is not solved
[by zero-shot learning], we do the following [description of the few-shot 
procedure]''. The question is, how does the system know that zero-shot
learning has not succeeded? As far as I can see the question is not answered
in the paper. Perhaps the system uses some legitimate method; e.g. the Codex 
system fails to produce executable code. However, if that were the criterion,
one would expect that some fraction of the time, zero-shot learning would
produce code that executes but is erroneous; and there is no suggestion
of that in the paper. What seems much more likely is that the system moves
to few shot learning {\em when zero-shot learning has produced an answer that 
is incorrect.} That is, the program is using the recorded correct answer to
guide its actions. That would be cheating\footnote{In the technical sense. I 
don't, certainly, mean to accuse the authors of deliberate malfeasance; 
merely of sloppiness and unintentional misrepresentation} and if that is the
case, then all of the results relative to few-shot learning must be thrown
out, or at least interpreted with a very large asterisk.

Finally, in one of the examples in the paper, the code is nonsense
though it gives the right answer. The question is ``Determine whether the
alternating series converge or diverge [sic]: 
$\sum_{n=1}^{\infty} (-1)^{n+1}/n^2$.'' 
The Python code produced by Codex is

\begin{verbatim}
from sympy import Sum, Symbol, oo, limit, init_printing

init_printing()

n = Symbol('n')

s = Sum(((-1)**(n+1))/n**2, (n, 1, oo))

limit(s,n,oo)

"""
The series converges.
"""
\end{verbatim}

If you actually run this, the result of the call to ``limit'' is the symbolic
expression $\sum_{n=1}^{\infty} (-1)^{n+1}/n^2$.  But having generated this
expression, the code proceeds to ignore it and simply print out
``The series converges''. (The example
that it is using for guidance in few shot learning has the same error.)

Even with all this borne in mind, the program remains impressive. My real 
complaint here is not about the program but about the paper. The paper has 
eighteen authors, of whom eleven are affiliated with M.I.T., four with 
Columbia, two with Harvard, and one with Waterloo. Presumably it was read 
by three reviewers for PNAS. How did all these sloppy errors get past all
these readers?

\subsection*{References}
Brown, Tom et al. (2020). ``Language models are few-shot learners.'' 
{\em Advances in Neural Information Processing Systems,} {\bf 33}: 1877-1901.

Chen, Mark et al. (2021). ``Evaluating large language models trained on code.'' 
arXiv preprint arXiv:2107.03374.

Drori, Iddo et al. (2022) 
``A neural network solves, explains, and generates university
math problems by program synthesis and few-shot learning at
human level.'' {\em Proceedings of the National Academy of Sciences (PNAS)}
119(32), p.e2123433119.

Hendrycks, Dan, et al. (2021).
"Measuring mathematical problem solving with the math dataset." 
arXiv preprint arXiv:2103.03874 

\end{document}